\documentclass[conference]{IEEEtran}
\usepackage{graphicx}
\usepackage{epstopdf}
\usepackage[bottom]{footmisc}
\usepackage{url}
\graphicspath{{./Figures/}}
\begin{document}
\title{Multimodal deep learning approach for joint EEG-EMG data compression and classification}

\author{\IEEEauthorblockN{Ahmed Ben Said, Amr Mohamed, Tarek Elfouly}
\IEEEauthorblockA{Computer Science\\ and Engineering Department\\
Qatar Univrsity\\
2713, Doha, Qatar\\
Email: \{abensaid, amrm, tarekfouly\}@qu.edu.qa}
\and
\IEEEauthorblockN{Khaled Harras}
\IEEEauthorblockA{Computer Science Department\\
Carnegie Melon university-Qatar\\
24866, Doha, Qatar\\
Email: kharras@qatar.cmu.edu}
\and
\IEEEauthorblockN{Z. Jane Wang}
\IEEEauthorblockA{Electrical and Computer \\Engineering Department\\
	University of British Columbia\\
	Vancouver, BC, Canada\\
	Email: zjanew@ece.ubc.ca}
}
\maketitle
\begin{abstract}
In this paper, we present a joint compression and classification approach of EEG and EMG signals using a deep learning approach. Specifically, we build our system based on the deep autoencoder architecture which is designed not only to extract discriminant features in the multimodal data representation but also to reconstruct the data from the latent representation using encoder-decoder layers. Since autoencoder can be seen as a compression approach, we extend it to handle multimodal data at the encoder layer, reconstructed and retrieved at the decoder layer. We show through experimental results, that exploiting both multimodal data intercorellation and intracorellation 1) Significantly reduces signal distortion particularly for high compression levels 2) Achieves better accuracy in classifying EEG and EMG signals recorded and labeled according to the sentiments of the volunteer.
\end{abstract}
\IEEEpeerreviewmaketitle
\section{Introduction}
Healthcare has always been considered as a strategic priority worldwide. The increasing number of elderly and chronic disease patients has made the physical contact between the caregiver and patients more and more difficult. Following the fast development of wireless technologies, the interoperability between healthcare entities and mobile has grown. The development of complex devices has stimulated the creation of many mobile health or `mHealth' applications and wearable devices for fitness tracker, sleep monitoring \cite{app}\ldots The mHealth industry is predicted to grow \$12 billions by 2018 \cite{study}.\newline
Motivated by the myriad of biomedical sensors, mobile phones and applications, the scientific communities have standardized the system that focuses on the acquisition of vital signs such as electroencephalogram (EEG) and Electromyogram (EMG) by body area sensor networks (BASN) under IEEE 802.15.6 \cite{ieee}. Thus, a typical mHealth BASN system consists of sensors collecting the data, a Personal Data Aggregator (PDA) and remote server. However, due to network limitation, data delivery through the network can be hindered. Consequently, we need to optimize every bit of data being sent. One of the possible pre-processing stages is to encode the data in the PDA i.e. mapping $x_i$ to compressed data $z_i$. At the server level, the received data $z_i$ is decoded i.e. mapped to $\hat{x_i}$  which approximates the original data $x_i$. Many successful algorithms have been proposed for time series compression. Srinvasan et al. \cite{lossless} designed a 2-D lossless EEG compression where the signal is arranged in 2-D matrix as a preprocessing. Compression is achieved through a two-stage coder composed of Set Partitioning In Hierarchical Trees (SPIHT) \cite{spiht} layer and Arithmetic Coding (RC) layer. Hussein et al. \cite{ramy} proposed a scalable and energy efficient EEG compression scheme based on Discret Wavelet Transform (DWT) and Compressive Sensing (CS) in wireless sensors. Several parameters have been considered to control the total energy consumption of the encoder and transmitter. The optimal configuration of these parameters is chosen based on optimization scheme where the total power consumption should not exceed a certain threshold. In \cite{cs_shukla}, authors applied CS technique for EEG signal compression. Since the multichannel EEG signals have common sparse support in the transform domain, they stack the sparse transform coefficients as columns of matrix. Thus, the recovery problem becomes row-sparse and solved through Bregman algorithm \cite{bregman}. Majumdar et al. \cite{low_rank} argued that CS is not efficient for EEG compression because there is no sparsifying basis that fulfills the requirements of incoherence and sparsity. Instead, authors formulated the problem as a rank deficiency problem solved by a Bregman-derived algorithm.\newline
Following the development of wireless BASN, vital signs data have become abundant. mHealth systems are now capable of collecting data from different modalities (EEG, EMG, etc). Although, they may seem totally different, these data can describe the same phenomena. For example, in case of schizophrenic person, when a stimulus is presented, a peak in the EEG registration is witnessed while the functional Magnetic Resonance  Imaging (fMRI) data shows activations in the temporal lobe and the middle anterior cingulate region \cite{fmri_eeg}. Thus, both modalities are very likely to be correlated. Each modality has its advantages and limitations but analyzing multiple modalities offers better understanding of the investigated phenomena. The aforementioned methods, although exhibit good performance, do not exploit the correlation among multiple modalities. \newline
Deep learning approach has emerged as one of the possible techniques to exploit the correlation of the data from multiple modalities. Ngiam et al. \cite{ngiam} proposed a multimodal deep learning approach for cross modality feature learning from video and speech data. Srivastava et al. \cite{nitish_m_dbn} built a multimodal deep belief network \cite{dbn} to learn multimodal representation from image and text data for image annotation and retrieval tasks. In \cite{m_dbm}, authors designed a deep Boltzmann machine \cite{dbm} based architecture to extract a meaningful representation from multimodal data for classification and information retrieval task. Liu et al. \cite{m_video} proposed a multimodal autoencoder \cite{science} approach for video classification based on audio, image and text data where the intra-modality semantic for each data is separately learning by a stacked autoencoder. Next, the learned features are concatenated and fed to another deep autoencoder with a softmax layer for classification.\newline
Few research attempts have addressed the possible application of autoencoder  for biomedical and mHealth applications. In \cite{Ollivier2014}, Yann Ollivier proved that there is a strong relationship between minimizing the code length of the data and minimizing reconstruction error that an autoencoder seeks.  Tan et al. \cite{mammogram} used a stacked autoencoder for mammogram image compression. Training is conducted on image patches instead of the whole images. In \cite{ecg_ae}, authors applied the autoencoder for Electrocardiogram (ECG) compression. Comparison results with various classic compression methods showed that this special type network is reliable for signal compression. However, the problem of multimodal data compression in context of mHealth is still not well-investigated. \newline
We propose in this paper a multimodal approach for  data compression and feature learning. The encoding-decoding scheme can be achieved through a stack of autoencoders. Our approach exploits the intracorrelation as well as the intercorrelation among multiple modalities to achieve efficient compression and classification. 
\newline
The rest of the paper is organized as follows: in section II, we present the autoencoder architecture. Section III is dedicated for presenting our multimodal approach for joint EEG and EMG data compression and classification. Experimental results are illustrated and discussed in section IV and we conclude in the last section.   
\section{Background}
\subsection{Autoencoder}
\begin{figure}[!b]
\centering
\includegraphics[scale=0.55]{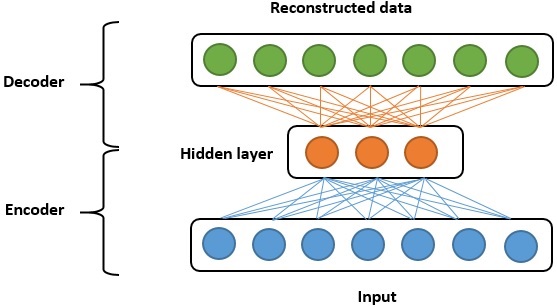}
\caption{Autoencoder: the encoder maps the data to hidden representation. The decoder maps the encoded features to reconstruct the data}
\label{autoencoder}
\end{figure}

An autoencoder, illustrated in Fig. \ref{autoencoder},  is a special type of neural network consisting of three layers. The data are first fed into the input layer, propagated to a second layer called the hidden or bottleneck layer and then reconstructed at a third layer called the reconstruction layer.
\newline
The encoder transforms the set of data vectors $x \in R^X$ into hidden representation  $h \in R^H$ via an activation function $f$:
\begin{equation}
h=f(Wx+b)
\end{equation}
The decoder transforms back the hidden representation $h$ to reconstruction data $r \in R^X$ via an activation function $g$:
\begin{equation}
r=g(W^{'}h+b^{'})
\end{equation} 
The Parameters $W \in R^{X \times H}$ and $W^{'} \in R^{H \times X }$ are called weight matrices. $b \in R^H$ and $b^{'} \in R^X$ are called the bias vectors. $f$ and $g$ are typically hyperbolic function $tanh(x)=\big(e^x-e^{-x}\big)/\big(e^x+e^{-x}\big)$ or sigmoid function $sigmoid(x)=1/\big( 1+e^{-x}\big)$. In practice, we use tight weight configuration i.e. $W^{'}=W^T$. Autoencoder seeks the optimal set of parameters $\Theta=\{W, b, b^{'}\}$ that minimizes the reconstruction error $J_{\Theta}(x,r)$. This error is generally the Squared Euclidean distance $ L(x,r)=|| x-r ||^2$ or cross-entropy loss $L(x,r)=-\sum^X_{i=1}x_i log(r_i)+(1-x_i)log(1-r_i)$. When using affine activation function and squared error loss, autoencoder essentially performs Principal Component Analysis (PCA) \footnote{It will find the same subspace as PCA but the projection direction does not essentially correspond to the principal components directions}.  Minimization is generally carried out  via gradient descent algorithm. In other words, the purpose of this minimization is to obtain $r \approx x$ i.e. an approximation to the identity function. But, by constraining the system by  limiting the number of hidden units at the hidden layer, we are forcing the system to learn a compressed version of the data. Furthermore, to prevent overfitting, i.e. just learning the identity function, another constraint is often added: a weight decay term that regularizes $J_{\Theta}(x,r)$. Then, we have:
\begin{equation}
J_{\Theta}(x,r)=L(x,r)+ \lambda||W||^2_2
\end{equation}
Where $\lambda$ is the decay parameter that controls the amount of regularization. 
\subsection{Stacked autoencoder}
\begin{figure}[!t]
\centering
\includegraphics[scale=0.55]{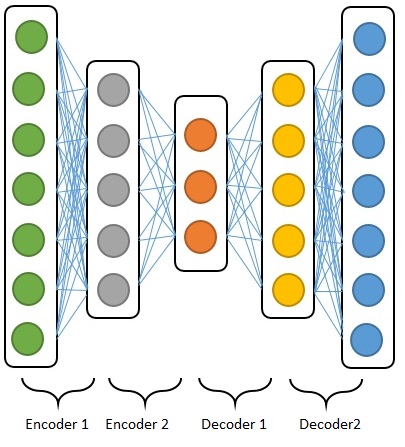}
\caption{Stacked autoencoder: the output of each layer is the input of the next layer}
\label{sae}
\end{figure}

Stacked autoencoder (SAE), illustrated in Fig \ref{sae}, is a neural network which consists of multiple layers of autoencoders. The output of each layer is fed to the next layer. SAE is trained via a greedy layer-wise training \cite{greedy}. Specifically, it is done one layer at a time. At each layer, we consider the autoencoder composed of the current layer and its previous one which is the output of the previous layer. Once $N-1$ layers are trained, we can compute the output of th $N^{th}$ layer wired to it. This unsupervised stage is followed by a supervised fine-tuning of the parameters where a softmax layer is added on top of the SAE.  
\begin{figure*}[!t]
\centering
\includegraphics[scale=0.45]{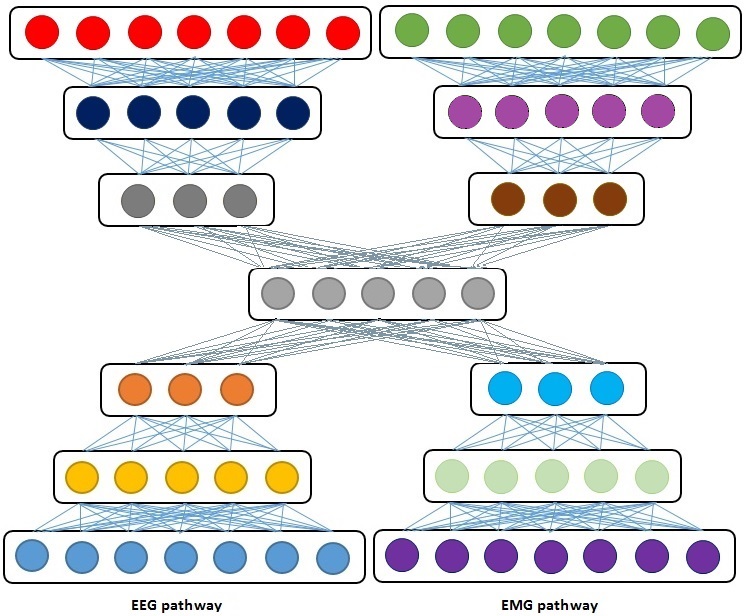}
\caption{Multimodal stacked autoencoder with EEG and EMG pathways. The input of each pathway is the set of vectors of each data with the number of units corresponding to the data dimension}
\label{m_sae}
\end{figure*}
\section{Multimodal autoencoder for EEG-EMG compression and classification}
Fig. \ref{m_sae} exhibits the multimodal autoencoder architecture. It consists of two pathways for EEG and EMG. Each pathway represents a unimodal stacked autoencoder dedicated to learn the intra-modality correlation of the data while the joint layer merge the higher level features.  

\subsection{Unimodal data pre-training}
SAE is applied separately for each modality, we use the sigmoid activation function and the Squared Euclidean distance as loss function regularized by a weight decay term. We apply also tied weight configuration. The output of the $i^{th}$ layer is obtained as follows:
\begin{equation}
\begin{tabular}{c c}
\(z_1=sigmoid(W_1 x_1+b_1)\) & \(i=1\)\\
\(z_i=sigmoid(W_i x_i+b_i)\) &  \(i=2..N\)
\end{tabular}
\end{equation} 
The SAE is trained using the greedy-layer wise training approach where we feed the latent representation of the autoencoder found below to the current layer. This deep architecture makes the system more scalable and efficient while progressively extracting  higher level features from the high dimensional data. 
\subsection{Deep multimodal learning}
The single modal pre-training does not involve inter-modality correlation which can contribute in better representation of the higher level features. It especially allows encoding the multiple modalities in a single shared representation obtained by the joint layer. The output of this layer encompasses the contribution of each modality in the code which represents the compressed data. The joint representation is obtained as follows:
\begin{equation}
z=\sum_{ i \in \{e,m\}}sigmoid\big(W_{N+1}^i z^i_{N+1}+b^i_{N+1}\big)
\end{equation}
Where $e$ and $m$ refer to EEG and EMG respectively. Furthermore, we train the multimodal autoencoder with an augmented noisy data where additional examples are added leading to samples with only one single modality. In practice, we add zeros values examples for one modality while keeping the original values for the other modality and vice-versa. Thus, one third of the training data is EEG only, another one third is EMG only and the rest has both EEG and EMG data. This strategy, inspired from Nigiam et al \cite{ngiam}, follows the denoising autoencoder paradigm \cite{dae} and is justified by twofold: 
\begin{itemize}
\item Correlation among multiple modalities is very likely to be non-linear.
\item This non-linearity often leads to hidden units being activated by one single modality.
\end{itemize}
Therefore, the original and corrupted inputs are propagated independently to the higher layers which are then trained progressively to reconstruct the clean presentation from both inputs.
\subsection{Fine-tuning}
The compressed data can be used for classification task, that is, to fine-tune the layers with respect to a supervised criterion by plugging the bottleneck layer to a softmax classifier \cite{autoencoder}:
\begin{equation}
\hat{p}=\frac{exp(Wy+b)}{\sum_{l=1}^L exp(W^l y+b^l)}
\end{equation}   
Where $\hat{p}$ is the predicted object label, $y$ represents the compressed data and $L$ is the number of classification labels. Therefore, the overall objective function to minimize is:
\begin{equation}
\Im(x,r,p,\hat{p})=J_{\Theta}(x,r)+ L(p,\hat{p})
\end{equation}
Where $p$ is the true label and $L(p,\hat{p})$ can be an entropic loss function.
\section{Experimental results}
mHealth systems acquire, process, store, secure and transport the medical data. Data delivery should be as efficient and optimized as possible in terms of energy consumption and bandwidth usage. A typical system consists of mHealth wearable device that senses vital signs. These data are collected by a PDA and should be transmitted to a remote server handled by a medical entity \cite{7552682}. At the server level, a multimodal autoencoder is already trained and the optimal configuration is already found. This configuration is also known by the PDA which should apply it on the collected data for compression. \newline 
We present in this section several experimental results where we compare our compression scheme with some state of the art compression methods. Furthermore, we compare our multimodal strategy with the unimodal one to highlight the importance of exploiting the intermodality correlation.

\subsection{Dataset}
We conduct our experiments on the DEAP dataset \cite{deap}. It consists of EEG, EMG and multiple physiological signals recorded from 32 participants during 63 seconds at 128 Hz. During experiments, volunteers watched 40 music videos and rate them on a scale from 1 to 9 with respect to four criteria: likeness (dislike, like), valence (ranges from unpleasant to pleasant), arousal (ranges from uninterested or bored to excited) and dominance (ranges from helpless and weak feelings to empowered feeling). \newline
Signals are segmented into 6 seconds segments, whitened and normalized between 0 and 1. For both EEG and EMG data, we have 23040 samples of 896 dimensionality. These data should then be divided into training and testing sets.
\subsection{Compression tasks}
\begin{figure}[h!]		
	\centering		
	\includegraphics[scale=0.22]{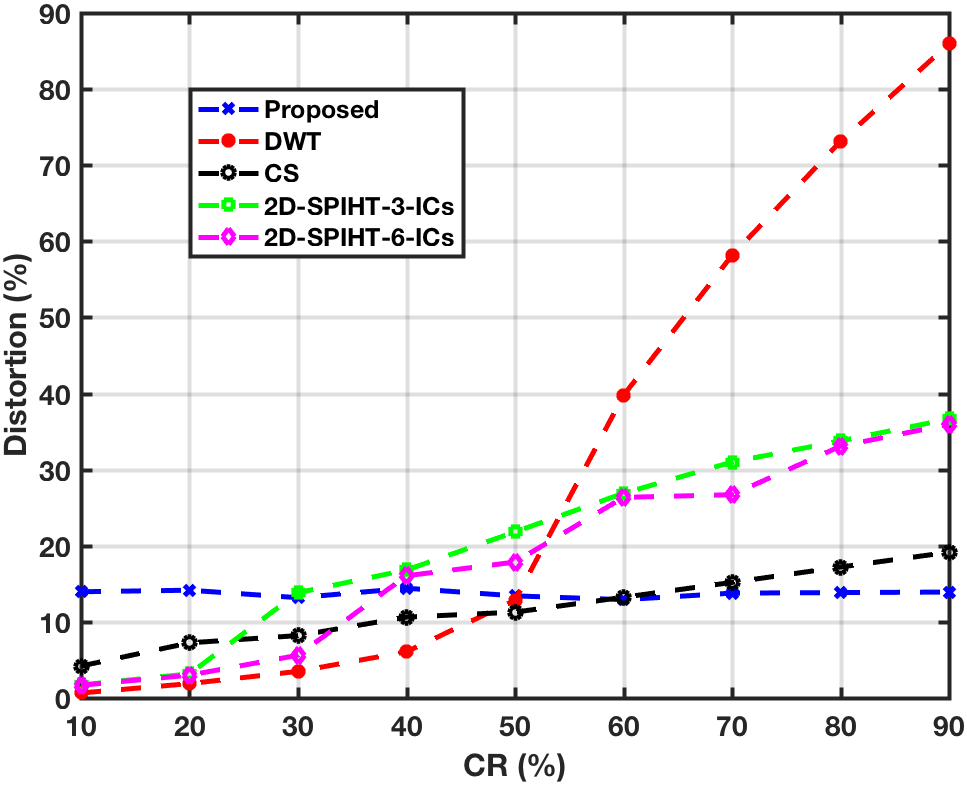}	
	\caption{EEG Distortion (\%) with respect to the Compression Ratio (\%). }	
	\label{distortion}		
\end{figure}

\begin{figure}[h!]	
	\centering
	\includegraphics[scale=0.22]{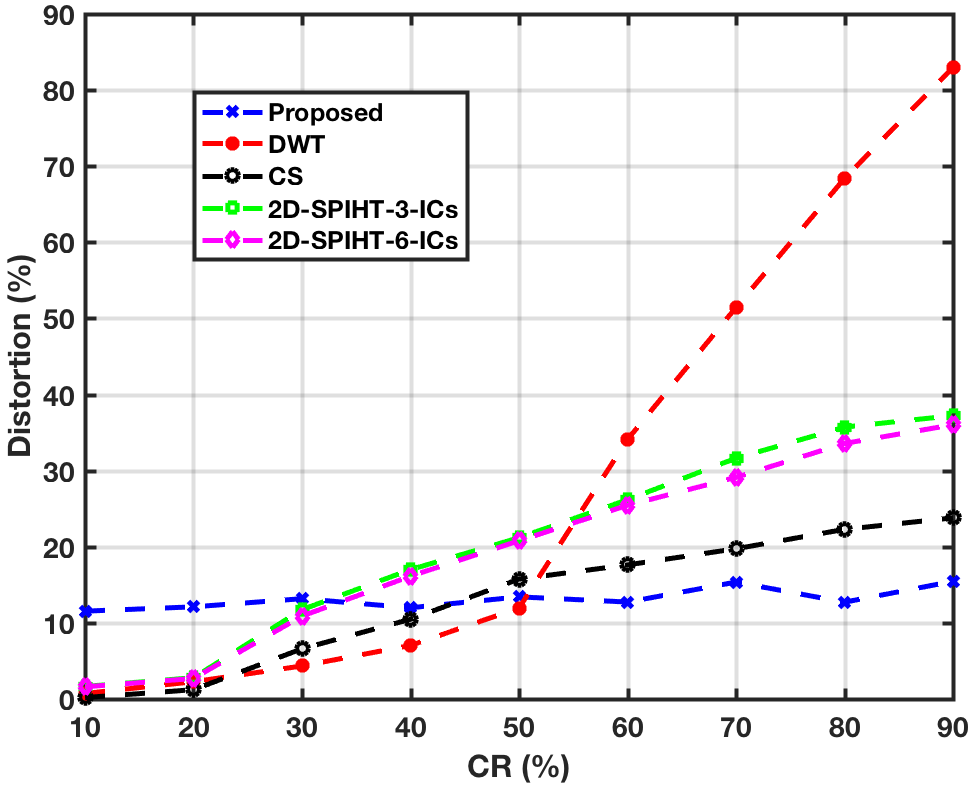}
	\caption{EMG Distortion (\%) with respect to the Compression Ratio (\%). }
	\label{distortion_emg}	
\end{figure}

\begin{figure}[b!]	
	\centering
	\includegraphics[scale=0.2]{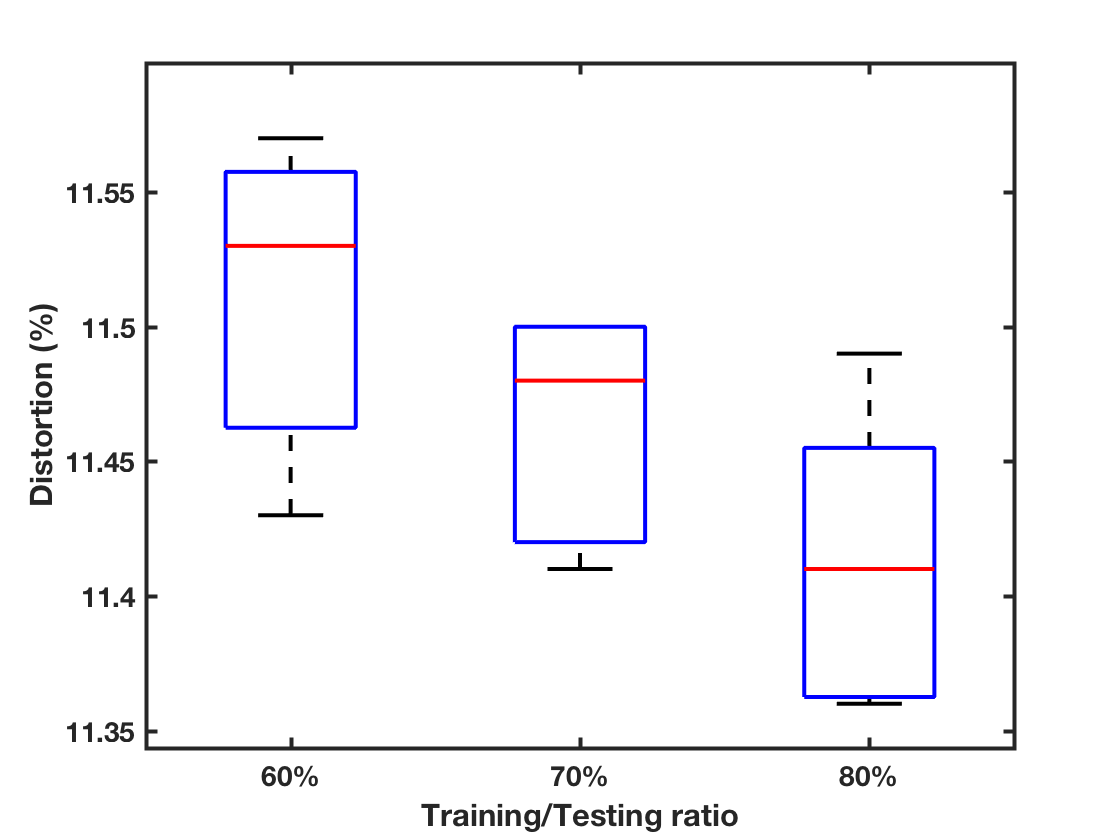}
	\caption{Whisker diagram of EEG data with various training/testing partitions}
	\label{eeg_boxplot}	
\end{figure}

\begin{figure}[t!]	
	\centering
	\includegraphics[scale=0.2]{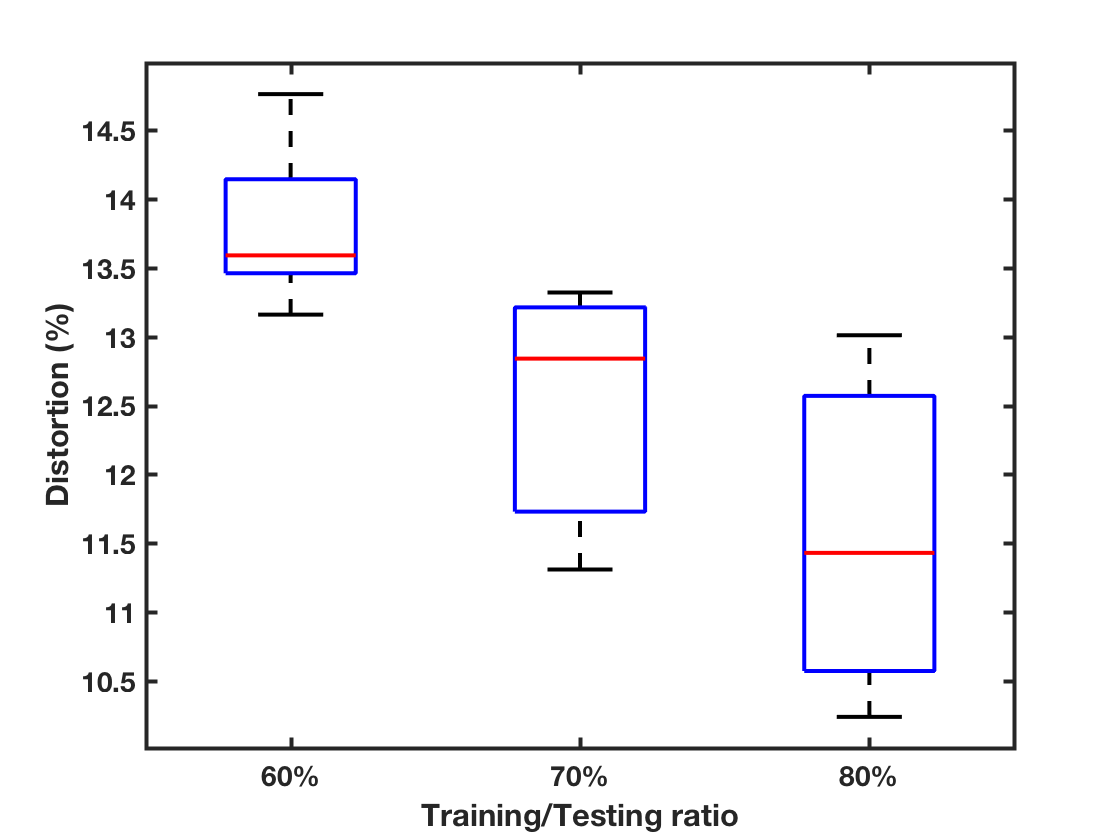}
	\caption{Whisker diagram of EMG data with various training/testing partitions}
	\label{emg_boxplot}	
\end{figure}

We compare our compression method with the Discrete Wavelet Transform (DWT) \cite{dwt}, Compressed Sensing (CS) \cite{cs} and the 2D compression approach which is based on SPIHT and FastICA \cite{fastica}. For the latter algorithm, we use two configurations of 3 and 6 independent components denoted 2D-SPIHT-3-ICs and 2D-SPIHT-6-ICs. We evaluate performance using compression ratio (CR) and residual distortion (D).
\begin{equation}
CR=\big(1-\frac{m}{n}\big)*100
\end{equation} 
\begin{equation}
D=\frac{||r-x||}{||x||}*100
\end{equation}
Where $m$ and $n$ are the length of the compressed and original signals (number of samples). $D$ is the percentage root-mean-square difference between the compressed and original signals. For each data pathway, we use a two-layer SAE. Table \ref{config} presents the numbers of hidden units for each layer of the multimodal autoencoder as well as the DWT thresholds and their corresponding CRs. We divide data to 50\% training and testing. \newline
Fig. \ref{distortion} and \ref{distortion_emg} exhibit distortion variation with respect to different CR values for EEG and EMG. The findings show that for higher compression ratios, the multimodal approach performs better than DWT and CS. For example, for CR=80\%, our approach is able to reconstruct EEG and EMG with distortions of 12\% and 13.85\% respectively while CS distorts EEG by 22\% and EMG by 17.21\%. With 2D-SPIHT-3-ICs, EEG and EMG distortions are 33.7\% and 35.7\% respectively while with 2D-SPIHT-3-ICs, EEG and EMG are distorted by 33\% and 33.5\%. DWT exhibits low performance with 68\% and 73.12\% for EEG and EMG respectively. Although DWT, CS and the 2D approach perform better for low compression levels, the proposed method presents stable performance for different compression levels. 
\begin{table}[t!]
\centering
\caption{ Multimodal autoencoder and DWT configuration and the resulting compression ratio}
\label{config}
\begin{tabular}{|c|c|c|}
\hline
Multimodal autoencoder & DWT threshold & CR (\%)\\ \hline 
896-806 & EEG: 0.025 ; EMG: 0.019 & 10 \\  \hline 
896-716 & EEG: 0.05  ; EMG: 0.04  & 20 \\  \hline
896-627 & EEG: 0.085 ; EMG: 0.06  & 30 \\  \hline 
896-537 & EEG: 0.13 ; EMG: 0.10   & 40 \\  \hline
896-448 & EEG: 0.29 ; EMG: 0.51   & 50 \\  \hline 
440-358 & EEG: 0.66 ; EMG: 0.64   & 60 \\  \hline
440-268 & EEG: 0.75 ; EMG: 0.69   & 70 \\  \hline 
440-179 & EEG: 0.83 ; EMG: 0.74   & 80 \\  \hline
380-89  & EEG: 0.92 ; EMG: 0.78   & 90 \\  \hline 
\end{tabular}
\end{table} 
 This can be explained by the capacity of the underlying architecture to exploit the statistics of the data to achieve better compression.\newline
 We further examine the effect of training/testing data partition on the compression results. Fig. \ref{eeg_boxplot} and Fig. \ref{emg_boxplot} illustrate the whisker diagrams for EEG and EMG signals respectively. We can clearly deduce that more training data result in less distortion. This confirms a known deep learning rule of thumb stating the more training data we have, the better the results are.
 \subsection{Classification task}
\begin{figure}[h!]	
	\centering
	\includegraphics[scale=0.5]{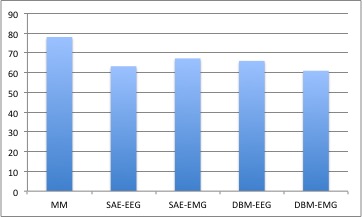}
	\caption{Classification accuracy with respect to the dominance label: the multimodal approach (MM) achieves the best performance with 78.1\%}
	\label{accuracy_dominance}	
\end{figure}

\begin{figure}[h!]	
	\centering
	\includegraphics[scale=0.5]{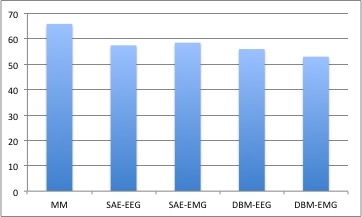}
	\caption{Classification accuracy with respect to the arousal label: the multimodal approach (MM) achieves the best performance with 65.9\%}
	\label{accuracy_arousal}	
\end{figure}

 The objective of this experiment is to demonstrate the importance of the multimodal approach. We conduct binary classification of the EEG and EMG with respect to two of the four labeling possibilities: dominance and arousal. We follow the same approach as in \cite{deap}: video ratings are thresholded into two classes. On the scale of 1 to 9, we simply place the threshold in the middle. We compare our approach with two-layer SAE and Deep Boltzmann Machine (DBM) architectures \cite{dbm} with the softmax classifier on top of them. For SAE, we use the sigmoid activation function. We choose 75\% training-testing partition. Figures \ref{accuracy_dominance} and \ref{accuracy_arousal} illustrate the classification results with respect to the dominance and arousal respectively. By exploiting the inter-modality correlation, the proposed approach achieves the best results with 78.1\%. The single-modality approaches are less accurate. \newline 
 These findings confirm that, when available, multiple modalities can offer better understanding of the underlying phenomena even if the data exhibit different characteristics. 
 \subsection{Discussion}
 In a typical mHealth system, a client-server architecture is the common choice where the system relies on the available networks to deliver the data. In general, the healthcare giver generally relies on multiple vital signs for an accurate diagnosis. The proposed approach is flexible in the sense that, if an additional modality is collected by the PDA via a wearable device, can be easily incorporated in the architecture presented in Fig. \ref{m_sae}, compressed and classified. The deep neural network can be trained offline. Once it achieves good performance, the optimal configuration (weights and biases) is applied at the client side for efficient data delivery. However, it is worth-noting that our approach it less efficient for low compression ratio. 
\section{Conclusion}
We have presented a deep learning approach for multimodal data compression and classification. Our strategy focuses on exploiting the inter and intra correlation among multiple modalities to enhance the compression and classification of data in context of mHealth application. \newline 
The core of the proposed method is based on the classic autoencoder which has been originally designed for encoding-decoding data. For each modality presented, we dedicate a stacked autoencoder to extract high level abstraction of the data by modeling the intra-correlation. A joint layer is added on top of each encoding part of the stacked autoencoders to model data intercorrelation.\newline
We have conducted compression and classification experiments. Comparison with DWT and CS have shown that our approach performs better with high compression ratio. We have also demonstrated the effectiveness of the multimodal approach for classification of EEG and EMG. Comparison with some unimodal algorithms e.g. Deep Botzmann Machines and stacked autoencoders shows that the multimodal autoencoder leads to better classification accuracy. 
\newline
In future work, we will investigate the possible application of Convolutional Neural Network. Furthermore, we intend to make the autoencoder-based compression scheme adaptive by including the network resource  in the choice of the neural network architecture.
\section*{Acknowledgment}
This publication was made possible by NPRP grant \#7-‐684-‐1‐-127 from the Qatar National Research Fund (a member of Qatar Foundation). The statements made herein are solely the responsibility of the authors.

\bibliographystyle{IEEEtran}

\end{document}